\documentclass{article}
\usepackage{microtype}
\usepackage{graphicx}
\usepackage{subfigure}
\usepackage{booktabs} 
\usepackage[table]{xcolor}
\usepackage{array,hhline}
\usepackage{multirow}
\usepackage{diagbox}
\usepackage[toc,page]{appendix}
\usepackage{siunitx}
\usepackage{hyperref}

\usepackage[accepted]{icml2018}

\icmltitlerunning{Privacy Aware Offloading of Deep Neural Networks}

\usepackage[compact]{titlesec}
\titlespacing{\section}{0pt}{*0}{*0}
\titlespacing{\subsection}{0pt}{*0}{*0}
\titlespacing{\subsubsection}{0pt}{*0}{*0}

\def\rot{\rotatebox}

\begin{document}

\twocolumn[
\icmltitle{Privacy Aware Offloading of Deep Neural Networks}



\icmlsetsymbol{equal}{*}

\begin{icmlauthorlist}
\icmlauthor{Sam Leroux}{to}
\icmlauthor{Tim Verbelen}{to}
\icmlauthor{Pieter Simoens}{to}
\icmlauthor{Bart Dhoedt}{to}

\end{icmlauthorlist}

\icmlaffiliation{to}{Ghent University - imec, IDLab, Department of Information Technology}

\icmlcorrespondingauthor{Sam Leroux}{sam.leroux@ugent.be}

\icmlkeywords{Machine Learning, ICML}

\vskip 0.3in
]



\printAffiliationsAndNotice{}

\begin{abstract}
Deep neural networks require large amounts of resources which makes them hard to use on resource constrained devices such as Internet-of-things devices. Offloading the computations to the cloud can circumvent these constraints but introduces a privacy risk since the operator of the cloud is not necessarily trustworthy. We propose a technique that obfuscates the data before sending it to the remote computation node. The obfuscated data is unintelligible for a human eavesdropper but can still be classified with a high accuracy by a neural network trained on unobfuscated images.
\end{abstract}

\vspace{-0.5cm}
\section{Introduction}
Remote processing of neural networks in the cloud is not without risk. Traditional encryption techniques can protect the data while sending it to the cloud but the unencrypted data is needed by the computation node to evaluate the neural network. The operator of the computing node can not necessarily be trusted and has access to the raw data of the users. An even greater risk is the compromise of the node by a third party. Recent security breaches such as the leak of personal images stored in Apple iCloud or the abuse of personal data shared on Facebook for political goals have raised public awareness of privacy and security risks.


In this paper we present a technique to obfuscate the data before sending it to the cloud. The obfuscation routine renders the data unintelligible for a human eavesdropper while still retaining enough structure to allow a correct classification by the neural network. We focus on image classification using deep neural networks (DNNs) since this is arguably one of the most common use cases for DNNs but this technique could be applied to other application domains
as well.

Previous approaches to protect the privacy of users in computer vision tasks include extreme downsampling  \cite{ryoo2017privacy}, \cite{chen2017semi} and blurring or scrambling \cite{manolakosprivacy} of the inputs. These are hand-crafted heuristics that are able to remove privacy sensitive details but they also have a large penalty on the classification accuracy. The most similar approach to our work is \cite{ren2018learning} where the authors introduce a trainable model that modifies video frames to obfuscate each person's face with minimal effect on action detection performance. The biggest difference is that they train the classification network together with the obfuscation network. The classification network will therefore only work together with the obfuscation network. In contrast, we use pretrained classification networks that were trained on unobfuscated images. We then train an obfuscation network to transform images in order to make them unintelligible for humans while still allowing for a high classification accuracy with the pretrained classification network. We also obfuscate the full image instead of only specific parts of the human face.

\section{Architecture}
Our approach builds upon two recent discoveries in deep learning: Adversarial inputs and Generative Adversarial Networks (GANs). Adversarial inputs \cite{goodfellow2014explaining} are special input samples that have been carefully tweaked to fool neural networks. They are created by making tiny changes to real inputs such that the real and the perturbed versions are indistinguishable to human observers yet the model consistently misclassifies the perturbed input with high confidence. In this paper we are however interested  in the exact opposite behaviour, we want to transform images in order to make them unintelligible for human observers yet the neural network should still be able to correctly classify them. 

Generative Adversarial Networks (GANs) \cite{goodfellow2014generative} are models that can learn to generate artificial datapoints that follow the same distribution as real datapoints. The model consists of two networks, the generator and the discriminator competing against each other. The task of the generator is to generate new datapoints based on random input. The discriminator tries to distinguish between real data points and generated data. By training both networks together the generator will eventually be able to generate realistically looking datapoints. 

\begin{figure}[H]
\caption{Overview of our architecture.}
\label{fig:architecture}
\centering
\includegraphics[width=0.5\textwidth]{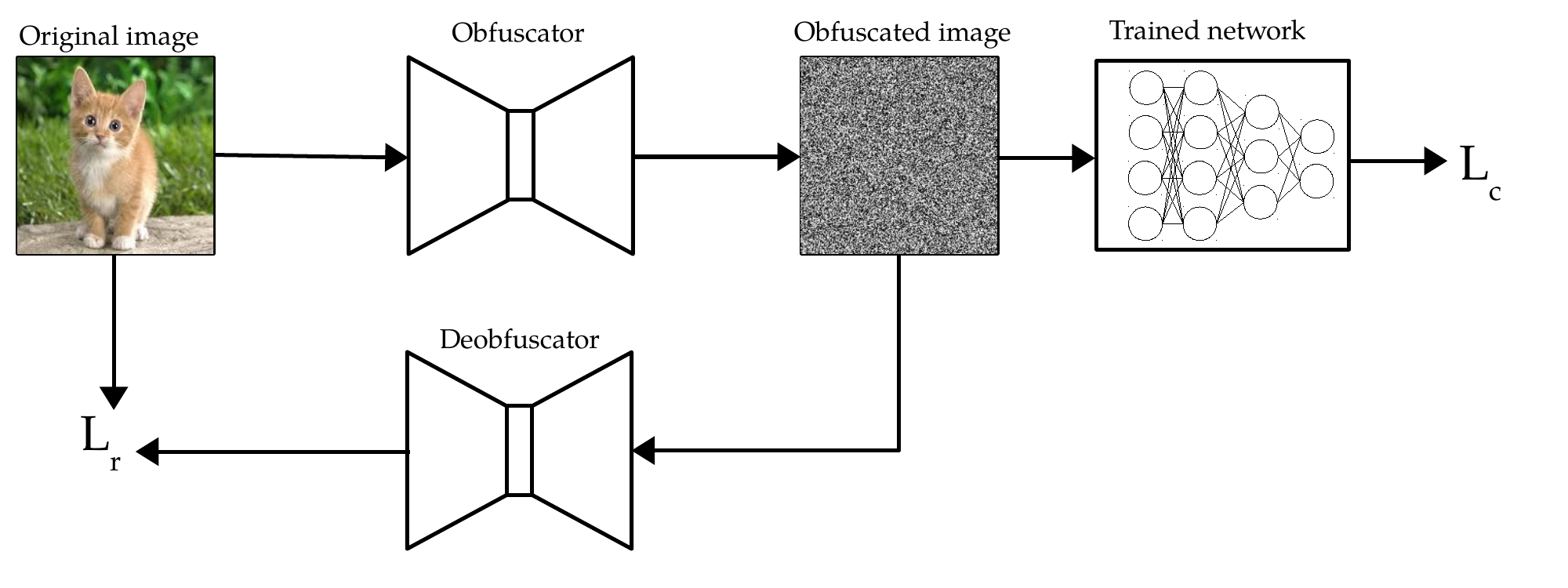}
\end{figure}

Our proposed architecture is shown in figure \ref{fig:architecture}. It consists of three deep neural networks. The pretrained network on the right is a network trained for image classification on normal, unobfuscated images. We do not modify the weights of this network. The obfuscator and the deobfuscator are two autoencoder-like networks. The obfuscator takes the original image as input and generates an obfuscated version that is then fed into the pretrained classification network.  The deobfuscator tries to reconstruct the original image based solely on the obfuscated version. The final goal is to train an obfuscator network than can transform the image in order that the classifying network is still able to recognize the object without the deobfuscator being able to reconstruct the original input. We introduce two loss terms to train the architecture. ${L_c}$ is the crossentropy classification loss that is commonly used in classification problems. ${L_r}$ is the reconstruction loss that measures the euclidean distance between the original image and the reconstructed version. The obfuscator is jointly trained to minimize the classification loss and to maximize the reconstruction loss. The deobfuscator is solely trained to minimize the reconstruction loss. Both networks are trained at the same time.

The premise of our approach is to offload the computationally costly classification network to the cloud and to do a local obfuscation step to protect the privacy of the user. It is therefore crucial that the obfuscator network is as small as possible. We use a MobileNet inspired architecture \cite{howard2017mobilenets} with depthwise seperable convolutions to reduce the computational cost. Details of the architecture and training routine can be found in Appendix \ref{app:one}.

\vspace{-0.5cm}
\vspace{-1cm}

\section{Experiments}
All our experiments were implemented in PyTorch \cite{paszke2017automatic} 
We used the CIFAR10 dataset \cite{krizhevsky2009learning} for all these experiments.

Table \ref{tbl:accuracy} shows the accuracy of the classification models on original and on obfuscated images. We find similar results for the different architectures where the accuracy drops by ~5\%. We argue that this is a reasonable price to pay for the added privacy. We also show the overhead of the obfuscation network relative to the classification network both in terms of FLOPS and number of parameters. Table \ref{tbl:images} shows some more qualitative results. We show the original images, the obfuscated and the deobfuscated versions. The classification network was trained on the original images but is also able to classify the obfuscated versions. The deobfuscated versions were included to prove that it is indeed impossible to retrieve the original images from the obfuscated versions. These images show that there is still information on the background color and the location of the object encoded in the obfuscated image but all details are lost.

One disadvantage with our proposed approach is that we need to backpropagate through the classification network to train the obfuscation network. This means that our technique does not treat the classification model as a truly black box since we need the weights of the network which might be unavailable. In our last experiment we examine how transferable the obfuscator networks are. We train the obfuscator network with one classification network and test it with another. The results are shown in Table \ref{tbl:transfer}. There is a large drop in accuracy but surprisingly the accuracy does not drop to the random level for most combinations. This suggests that the obfuscator network can learn a transformation that is not completely overfitted to one classification network but that captures some universal features that are used by different classification networks.

\vspace{-0.5cm}
\begin{table}[H]
\caption{Classification accuracies for plain and obfuscated images. The overhead column shows the cost (in terms of FLOPS and parameters) of the obfuscator network relative to the classification network. Absolute measurements are included in Appendix \ref{app:one}.}
\label{tbl:accuracy}
\begin{center}
\begin{scriptsize}
\begin{sc}
\begin{tabular}{lcc|cc}
\toprule
&\multicolumn{2}{c}{Accuracy}&\multicolumn{2}{c}{Overhead}\\
\midrule
Architecture & Plain & Obfusc. & Flops & Param.\\
\midrule
VGG19 		& 93.4\%  	 & 89.3\%		& 6.7\%		& 1.6\%	\\
ResNet18    & 94.8\% 	 & 89.8\%		& 4.8\%		& 2.9\%\\
ResNet50 	& 95.1\%		 & 90.2\% 		& 2.1\%		& 1.4\%\\
GoogleNet 	& 95.2\%	 	 & 90.5\% 		& 1.7\% 		& 5.3\% \\
\bottomrule
\end{tabular}
\end{sc}
\end{scriptsize}
\end{center}
\end{table}

\vspace{-1cm}
\begin{table}[H]
\caption{Original, obfuscated (output of the obfuscator) and reconstructed images (output of the deobfuscator). More examples are shown in Appendix \ref{app:two}}
\vspace{-0.35cm}
\label{tbl:images}
\begin{center}
\begin{scriptsize}
\begin{sc}
\begin{tabular}{ccc|ccc}
\toprule
\rot{45}{Original} & \rot{45}{Obfusc.} & \rot{45}{Reconstr.} & \rot{45}{Original} & \rot{45}{Obfusc.} & \rot{45}{Reconstr.}\\
\midrule
\includegraphics{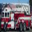} & \includegraphics{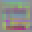} & \includegraphics{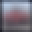} & \includegraphics{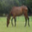} & \includegraphics{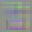} & \includegraphics{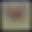}\\

\includegraphics{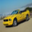} & \includegraphics{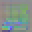} & \includegraphics{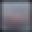} & \includegraphics{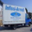} & \includegraphics{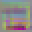} & \includegraphics{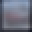}\\

\bottomrule
\end{tabular}
\end{sc}
\end{scriptsize}
\end{center}
\end{table}

\vspace{-1cm}
\begin{table}[H]
\caption{The classification accuracy when applying the obfuscator network to other networks than it was originally trained with. ResNet18\_1 and ResNet18\_2 are the same architecture, trained in the same way but from a different random initialisation.}
\vspace{-0.35cm}
\label{tbl:transfer}
\begin{center}
\begin{scriptsize}
\begin{sc}

\begin{tabular}{p{1.7cm}|p{0.9cm}p{0.9cm}p{0.9cm}p{0.9cm}p{0.9cm}cc}
\toprule
& \multicolumn{5}{c}{Tested on}\\
\cmidrule{2-6}
\rot{0}{Trained on} & \rot{45}{ResNet18\_1} & \rot{45}{ResNet18\_2} & \rot{45}{ResNet50} & \rot{45}{GoogleNet} & \rot{45}{VGG19}\\
\cmidrule{1-6}
 ResNet18\_1 & 89.8\% & 53.8\% & 36.0\% & 30.6\% & 47.4\%\\
ResNet18\_2 & 72.6\% & 90.0\% & 46.7\% & 34.7\% & 46.0\%\\
ResNet50 & 62.1\% & 54.9\% & 90.2\% & 29.1\% & 44.8\%\\
GoogleNet & 71.0\% & 74.0\% & 66.3\% & 90.5\% & 41.5\%\\
VGG19 & 25.8\% & 20.0\% & 20.2\% & 13.0\% & 89.3\%\\
\bottomrule
\end{tabular}
\end{sc}
\end{scriptsize}
\end{center}
\end{table}

\vspace{-0.7cm}
\section{Conclusion and future work}
We introduced a trainable obfuscation step that renders images unintelligible for humans but still allows a high classification accuracy by pretrained networks. Future work will focus on applying this technique to more complex datasets and on improving the transferability of the obfuscator networks between classification networks.

\newpage
\section*{Acknowledgements}
We gratefully acknowledge the support of NVIDIA Corporation with the donation of GPU hardware used in this research.

\bibliography{example_paper}
\bibliographystyle{icml2018}

\newpage

\begin{appendix}
\section{Network architecture and training details}
\label{app:one}
The obfuscator and deobfuscator networks are based on the MobileNet \cite{howard2017mobilenets} architecture. MobileNets use depthwise seperable convolutions to reduce the computational cost and the number of parameters. The basic building block is the ``Bottleneck module'' that uses 3 convolutional layers: a 1x1 pointwise convolution that expands the number of input channels by a factor of six, a 3x3 depthwise convolution that applies a single 3x3 filter to each channel and a 1x1 pointwise convolution that performs a linear combination of information in different channels to reduce the number of channels again. We use BatchNorm \cite{ioffe2015batch} and Leaky ReLU activations \cite{maas2013rectifier} for all layers. The obfuscator has 324518 parameters (1.2 MB if stored as 32 bit floating point) and requires \num{2.6e7} FLOPS. All our models were trained using the Adam optimizer \cite{kingma2014adam} with initial learning rate 0.001. We trained for 100 epochs and divided the learning rate by 10 every 30 epochs. We used horizontal flips of the training images to augment the dataset. The ``Upsample Bottleneck'' layers use 2D nearest neighbour upsampling to double the spatial size.

\begin{table}[H]
\vspace{-0.5cm}
\caption{Obfuscator and deobfuscator network architecture.}
\label{tbl:detail_arch}
\begin{center}
\begin{scriptsize}
\begin{sc}
\begin{tabular}{llrr}
\toprule
Input size & Module & Output Channels & Stride\\
\midrule
3 x 32 x 32 	& Conv2D  & 32	& 1	\\
32 x 16 x 16 	& Bottleneck  & 32	& 2	\\
64 x 16 x 16 	& Bottleneck  & 64	& 2	\\
128 x 4 x 4 	& Bottleneck  & 128	& 2	\\
128 x 4 x 4 	& Upsample Bottleneck  & 64	& 1	\\
64 x 8 x 8 	& Upsample Bottleneck  & 32	& 1	\\
32 x 16 x 16 	& Upsample Bottleneck  & 3	& 1	\\

\bottomrule
\end{tabular}
\end{sc}
\end{scriptsize}
\end{center}
\end{table}

\vspace{-0.7cm}
\section{Additional examples}
\label{app:two}
\begin{table}[H]
\vspace{-0.5cm}
\caption{Original, obfuscated (output of the obfuscator) and reconstructed images (output of the deobfuscator).}
\vspace{-0.35cm}
\begin{center}
\begin{scriptsize}
\begin{sc}
\begin{tabular}{ccc|ccc}
\toprule
\rot{45}{Original} & \rot{45}{Obfusc.} & \rot{45}{Reconstr.} & \rot{45}{Original} & \rot{45}{Obfusc.} & \rot{45}{Reconstr.}\\
\midrule
\includegraphics{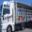} & \includegraphics{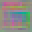} & \includegraphics{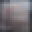} &

\includegraphics{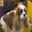} & \includegraphics{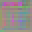} & \includegraphics{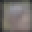}\\

\includegraphics{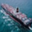} & \includegraphics{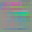} & \includegraphics{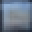} &

\includegraphics{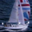} & \includegraphics{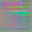} & \includegraphics{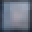}\\

\includegraphics{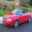} & \includegraphics{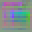} & \includegraphics{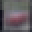} & 

\includegraphics{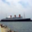} & \includegraphics{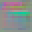} & \includegraphics{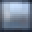}\\

\includegraphics{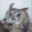} & \includegraphics{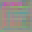} & \includegraphics{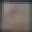} & 

\includegraphics{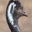} & \includegraphics{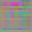} & \includegraphics{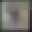}\\

\includegraphics{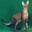} & \includegraphics{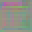} & \includegraphics{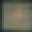} & 

\includegraphics{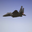} & \includegraphics{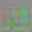} & \includegraphics{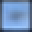}\\

\bottomrule
\end{tabular}
\end{sc}
\end{scriptsize}
\end{center}
\end{table}
\end{appendix}

\end{document}